\documentclass[conference,9pt]{IEEEtran}
\usepackage{newtxtext}
\DeclareFontFamily{OMS}{ntxtlf}{}
\DeclareFontShape{OMS}{ntxtlf}{m}{n}{<->ssub * cmsy/m/n}{}
\IEEEoverridecommandlockouts
\usepackage{cite}
\usepackage{amsmath,amssymb,amsfonts}
\usepackage{algorithmic}
\usepackage{graphicx}
\usepackage{textcomp}
\usepackage{xcolor}
\definecolor{lwblue}{HTML}{0057B8}
\newcommand{\hl}[1]{{\setlength{\fboxsep}{1pt}\colorbox{lwblue!15}{\vphantom{0}#1}}}
\usepackage{booktabs}
\usepackage[fixed]{fontawesome5}
\usepackage{hyperref}
\def\BibTeX{{\rm B\kern-.05em{\sc i\kern-.025em b}\kern-.08em
    T\kern-.1667em\lower.7ex\hbox{E}\kern-.125emX}}

\newcommand{\sys}{\texttt{LazyWeights}}
\AtBeginDocument{\setlength{\abovedisplayskip}{4pt plus 1pt minus 1pt}\setlength{\belowdisplayskip}{4pt plus 1pt minus 1pt}}
\newcommand{\lcpp}{\texttt{llama.cpp}}

\begin{document}

\title{LazyWeights: Decode-on-Demand Edge LLMs}

\author{\IEEEauthorblockN{
    Arnab Sanyal \IEEEauthorrefmark{2} \quad
    Gourav Datta \IEEEauthorrefmark{3} \quad
    Sandeep P. Chinchali \IEEEauthorrefmark{2}
    }
\IEEEauthorblockA{
    \begin{minipage}[t]{0.45\textwidth}
        \centering
        \href{https://case.edu/}{Case Western Reserve University}
    \end{minipage}
    \hfill
    \begin{minipage}[t]{0.45\textwidth}
        \centering
        \href{https://utexas.edu/}{The University of Texas at Austin}
    \end{minipage}
    \vspace{-0.5mm}
    }
}

\maketitle

\begin{abstract}
Memory determines which large language models (LLMs) an edge device can run and how fast it runs them: capacity limits a 4\,GB Jetson Nano, and bandwidth limits an 8\,GB Jetson Orin Nano Super. Deployed weight formats (GPTQ, AWQ, and \lcpp{}'s K- and I-quants) spend the same number of bits on every integer code of a matrix, although the codes are far from uniform. We present \sys{}, which keeps these formats, tunes GPTQ so the codes compress, Huffman-codes them losslessly in interleaved lanes, and decodes them on demand in a small unit between DRAM and L2, so GPU kernels run unchanged and DRAM moves only coded bytes. At 1\% and 3.5\% perplexity budgets, coded Phi-3-mini, Mistral-7B and SmolLM-1.7B are 5.7--31.4\% smaller than the best \lcpp{} format. A bit-exact 0.21\,mm$^2$ ASIC decoder unit (1\,GHz, 7\,nm) cuts Orin latency and energy per token by up to 36\% and 38\% (54\% and 44\% with prompt-lookup speculation). On the Nano, \sys{} needs 58--64\% less time and 53--57\% less energy per token than CPU \lcpp{}, and runs a streamed Mistral-7B 6.6$\times$ faster at 3.89 bits per weight. A Huffman decoder beats an rANS decoder in rate and area at equal throughput, and the unit runs bit-exact on a Zynq UltraScale+ FPGA (AMD ZCU102).\footnote{\href{https://arnabsanyal.com/preprint/lazyweights.html}{\faGlobe \texttt{website}}\quad\href{https://github.com/arnabsanyal/LazyWeights}{\faGithub \texttt{code}}\quad\href{mailto:sanyal@utexas.edu}{\faEnvelope\texttt{Corr: sanyal@utexas.edu}}}
\end{abstract}


\section{Introduction}\label{sec:intro}
Generating a token reads every weight of a large language model (LLM) once, so on an edge device the weight footprint decides both whether a model fits and how fast it runs. A 7B model needs 14\,GB in 16-bit floating point, more than most edge modules provide, and even at 4 bits every token moves gigabytes through a phone-class memory system. Post-training quantization has thus become the default: \lcpp{}~\cite{llamacpp} ships models as GGUF files in fixed-width integer formats with per-block scales, from 8 down to about 3 bits per weight.

These formats, and the quantizers that target them (GPTQ~\cite{gptq}, AWQ~\cite{awq}, and \lcpp{}'s K- and I-quants), share an implicit assumption: \emph{every code of a matrix costs the same number of bits}. This assumption wastes bits in two ways. First, weights are bell-shaped, so the inner codes of a 4-bit alphabet dominate and the code entropy lies well below 4 bits~\cite{physics33,memorize}. Second, many code assignments reach the same quality, so a quantizer that knows its codes will be entropy-coded can choose assignments that compress.

Lossless weight coding, from Deep Compression~\cite{deepcomp} to recent work on 16-bit LLM weights~\cite{dfloat11,zipnn,zipserv,huffllm} and on a quantizer's integers~\cite{sanyal_entrollm,entromcu}, takes the quantizer as given and mostly decodes in software. Yet a coded 4-bit model is worth deploying only if it beats the next-smaller \emph{fixed-width} format at equal quality, a comparison rarely made, and only if its codes can be expanded at DRAM rate on every token. Software decoders fall 15--20$\times$ short of that rate on an Orin Nano Super, making a token 2.5--3.2$\times$ slower than with the same codes at fixed width.

\begin{figure}[t]
\centering
\includegraphics[width=\columnwidth]{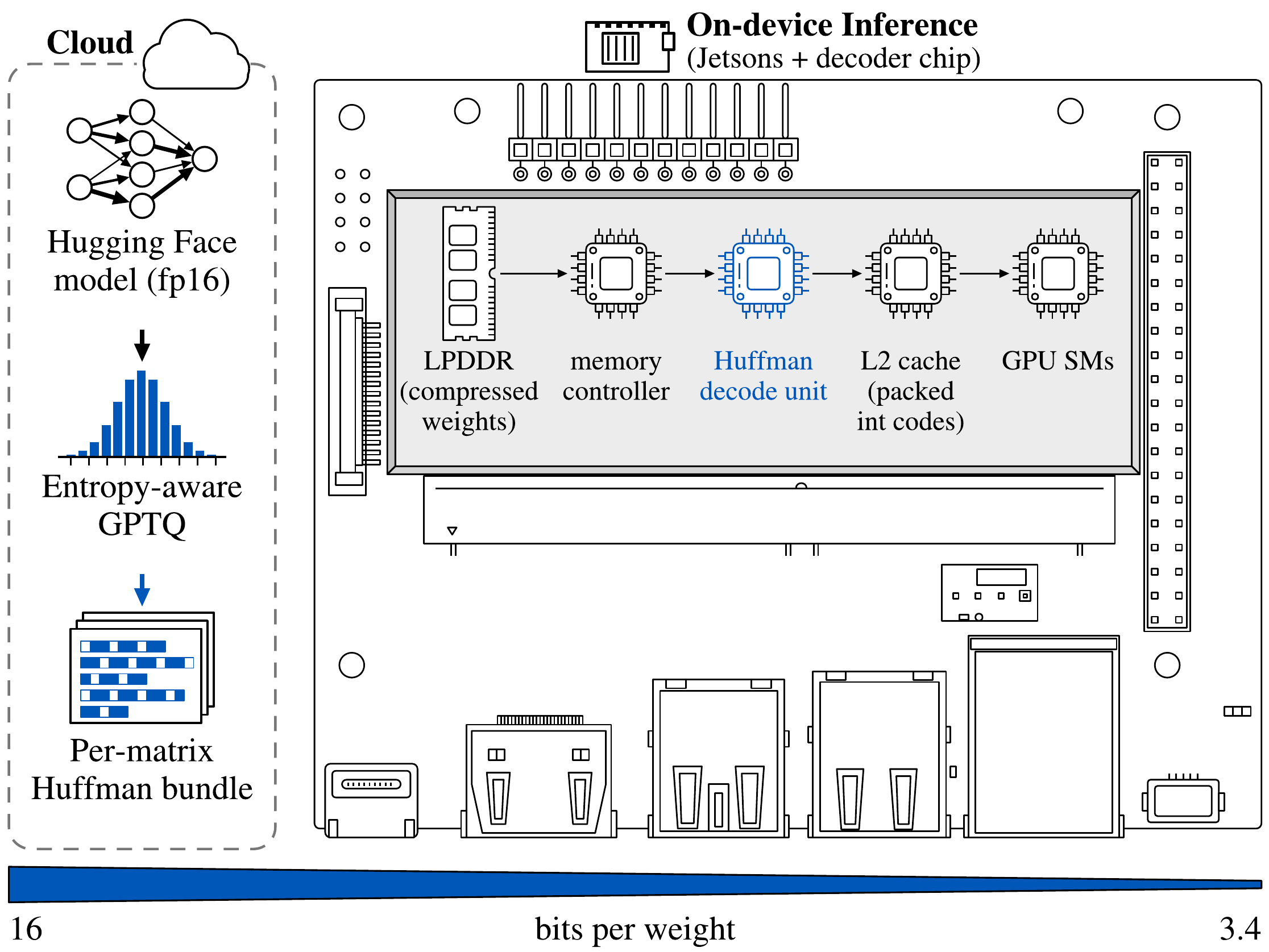}
\caption{Overview of \sys{}. In the cloud, entropy-aware quantization and a per-matrix Huffman bundle; on the device, a decode unit between the memory controller and L2 restores fixed-width codes. Without the unit, the GPU kernels decode in software inside GEMV (32 lanes), which serves as the reference. Bottom: bits per weight.}
\label{fig:overview}
\end{figure}

In this paper, we present \sys{}, which removes the fixed-width assumption end to end (Fig.~\ref{fig:overview}). \sys{} \emph{shapes} the codes within the deployed format, \emph{codes} them with a Huffman layout built for parallel decoding, and \emph{decodes} them in a small unit between DRAM and L2, where bytes are counted. Because the GPU still reads fixed-width codes, its kernels are untouched, the decoded weights are bit-identical to a valid GGUF file, and quality depends on the quantizer alone. On the bandwidth-bound Orin, fewer bytes then mean faster and cheaper tokens; on the instruction- and capacity-bound Nano, the unit removes software decoding, and the coded size decides whether Mistral-7B runs at all (Table~\ref{tab:platforms}).

The main contributions of this work are summarized as follows:
\begin{itemize}
\item \underline{\emph{Entropy-aware quantization in standard formats}}: GPTQ with shared sub-block steps and a wider step, emitting valid GGUF 5.7--31.4\% smaller than the best \lcpp{} format at $\le$1\% and $\le$3.5\% $\Delta$PPL.
\item \underline{\emph{A parallel-decodable Huffman bundle}}: one length-limited canonical code per matrix, 32 interleaved lane streams per tile, and pair symbols for 4-bit codes, within 1.2--2.1\% of the zeroth-order entropy with all overheads included.
\item \underline{\emph{A Huffman decode unit between DRAM and L2}}: sized from the bandwidth, 0.05\,mm$^2$ per 32-lane engine at 1\,GHz (ASAP7), bit-exact, and prototyped on an AMD ZCU102; never larger than rANS in rate, and smaller in area at equal throughput.
\item \underline{\emph{End-to-end evaluation in two edge regimes}}: latency and energy per token against \lcpp{} on a Jetson Orin Nano Super and a Jetson Nano, greedy and with prompt-lookup speculation.
\end{itemize}

\section{Related Work}\label{sec:related}
\textbf{Weight quantization.} GPTQ~\cite{gptq} quantizes a matrix column by column, feeding each column's Hessian-weighted error into the remaining columns, while AWQ~\cite{awq} rescales salient channels. Both target fixed-width formats, as do \lcpp{}'s~\cite{llamacpp} K- and I-quants (optionally with an importance matrix), in which edge models ship and which serve as our baseline. AQLM~\cite{aqlm}, QuIP\#~\cite{quipsharp} and QTIP~\cite{qtip} reach lower rates at equal quality with vector and trellis codebooks, but require their own formats and kernels. We instead keep the deployed scalar format and code its integers.

\textbf{Entropy-constrained quantization.} ECSQ adds a rate term to the distortion; Choi et al.~\cite{choi2017} applied Hessian-weighted ECSQ to network weights. Within GPTQ, structural changes to the grid matter far more than this term (\S\ref{sec:abl}).

\textbf{Lossless weight coding.} Deep Compression~\cite{deepcomp} Huffman-coded the cluster indices of pruned CNNs, and DeepCABAC~\cite{deepcabac} arithmetic-codes weights for storage and transmission. For LLMs, ZipNN~\cite{zipnn} compresses model files for storage and transfer, DFloat11~\cite{dfloat11} entropy-codes BF16 exponents and decodes them in a GPU kernel, ZipServ~\cite{zipserv} fuses a fixed-length bitmap encoding of BF16 into the Tensor-Core GEMM, and Huff-LLM~\cite{huffllm} Huffman-codes 16-bit weights end to end. All keep roughly 70\% of the BF16 size, about 11 bits per weight, well above what edge devices deploy. EntroLLM~\cite{sanyal_entrollm} codes tensor-level 8- and 4-bit integers and decodes them in parallel on a Jetson's CPU.

\textbf{Hardware entropy decoders.} Huffman decoding runs in FPGA Deflate cores~\cite{ledwon} and in parallel on GPUs~\cite{gpuhd}; ANS and rANS~\cite{duda} reach arithmetic-coding rates, but with a multiply in the state recurrence. APack~\cite{apack} places an arithmetic decoder next to a DNN accelerator's memory controller; we place a Huffman unit there in a GPU memory hierarchy, size it from the bandwidth, and compare it with rANS at equal throughput.

\textbf{Low-bit runtimes.} BitLLM~\cite{bitllm} accelerates fixed-width low-bit decoding on mobile GPUs (KV-cache layouts, packing, deferred type conversion); it reduces compute per code and we reduce bytes per code, so the two compose.

To our knowledge, \sys{} is the first design to \emph{code} the deployed integer format rather than merely store it, and to compare against the best fixed-width format at equal perplexity. Any quantizer that emits integer codes in a fixed format can feed its bundle and unit.

\section{Background and Motivation}\label{sec:bg}
\subsection{Decoding is memory-limited}\label{sec:platforms}
Each generated token runs one matrix--vector product (GEMV) per weight matrix, i.e., two FLOPs per weight, far below the arithmetic an edge GPU sustains per byte of bandwidth. Let $S$ be the weight bytes read per pass, $\beta$ the DRAM bandwidth, $t_\text{arith}$ the arithmetic time, and $S_\text{st}$ the bytes streamed from storage at bandwidth $\beta_\text{st}$ when memory cannot hold $S$. A weight pass then takes
\begin{equation}\label{eq:time}
t \approx \max\!\left(\frac{S}{\beta},\ t_\text{arith}\right) + \frac{S_\text{st}}{\beta_\text{st}}.
\end{equation}
Fewer bytes are faster only while $S/\beta$ dominates; when the model does not fit, the storage term dominates instead.

\begin{table}[t]
\caption{Edge platforms and the FPGA board. DRAM use: weight bytes per token over time per token, as a fraction of bandwidth, for our fixed-width kernels.}
\label{tab:platforms}
\centering\footnotesize
\setlength{\tabcolsep}{3pt}
\resizebox{\columnwidth}{!}{\begin{tabular}{@{}lcc@{}}
\toprule
 & Jetson Orin Nano Super & Jetson Nano \\
\midrule
Memory / bandwidth & 8\,GB, 102\,GB/s & 4\,GB, 22.5\,GB/s (meas.) \\
GPU & Ampere, 1024 cores & Maxwell, 128 cores \\
int8 \texttt{dp4a} / toolchain & yes / CUDA 12.6 & no / CUDA 10.2 \\
DRAM use, packed & 40--64\% & 34--35\% \\
Mistral-7B fits & yes & no (streamed, \S\ref{sec:stream}) \\
\midrule
Bound by & \textbf{memory bandwidth} & \textbf{instructions + capacity} \\
Compression buys & latency, energy & ability to run; the unit buys latency, energy \\
\midrule
\multicolumn{3}{@{}p{10.6cm}@{}}{\emph{Decoder prototype (\S\ref{sec:res-fpga}):} AMD ZCU102, Zynq UltraScale+ XCZU9EG ($-$2) with 274k LUTs, 548k flip-flops, 32.1\,Mb block RAM and 2,520 DSP slices; quad-core 1.2\,GHz Arm Cortex-A53 (Linux); 4\,GB 64-bit DDR4 (processor) and 512\,MB 16-bit DDR4 (fabric).} \\
\bottomrule
\end{tabular}}
\end{table}

On the Orin Nano Super, our fixed-width kernels match or beat \lcpp{} on the same GGUF and use 40--64\% of the bandwidth (Table~\ref{tab:platforms}), so time tracks bytes, and a unit that makes coded bytes the only bytes read turns compression into lower latency and energy. On the Nano, time tracks fp32 arithmetic and fewer bytes save little time; however, software Huffman decoding takes 54--69\% of a token, which the unit removes, and Mistral-7B does not fit, which puts the storage term of~\eqref{eq:time} in charge.

\subsection{The entropy of quantized codes}
A fixed-width format stores each code at the width of its alphabet, whereas an entropy coder stores it near the code entropy $H(q)$. For weights with density $p(w)$ quantized with step $\Delta$, at high resolution,
\begin{equation}\label{eq:ent}
H(q)\approx h(w)-\log_2\Delta,\qquad h(w)=-\!\int p(w)\log_2 p(w)\,dw,
\end{equation}
so scaling the step by $c$ saves about $\log_2 c$ bits per weight: 0.32 at $\times$1.25 and 0.58 at $\times$1.5. At 4 bits, far from that regime, the measured saving is smaller (0.19--0.20\,b/w at $\times$1.25 on Phi-3, 0.46 at $\times$1.5 on Mistral), but the lever is the same. Conversely, a quantizer that spreads codes evenly, as per-sub-block min-max fitting does, pushes $H(q)$ toward the full width. The saving of the coder is thus decided by the quantizer.

\section{The LazyWeights Design}\label{sec:method}
\subsection{Problem statement}
Let $\mathcal F$ be \lcpp{}'s formats (blocks, scales and code alphabet; e.g., IQ4\_XS, Q5\_K, Q6\_K) and $\theta$ the quantizer settings. Let $b(f,\theta)$ be the coded bits per weight, counting scales, code tables, lane offsets and embeddings, and $\Delta\mathrm{PPL}(f,\theta)$ the WikiText-2 perplexity increase over fp16. For a budget $\varepsilon$, we seek
\begin{equation}\label{eq:obj}
(f^*,\theta^\star)=\arg\min_{f\in\mathcal F,\ \theta}\ b(f,\theta)\quad\text{s.t.}\quad\Delta\mathrm{PPL}(f,\theta)\le\varepsilon,
\end{equation}
and compare against the \emph{budget baseline}: the smallest \lcpp{} file, with or without an importance matrix, within the same budget. $\varepsilon=1\%$ is near-lossless, the traditional role of 8 bits, and $3.5\%$ is a small, usable loss within which every model has a \lcpp{} point. Perplexity selects; downstream tasks confirm. Because the output must decode losslessly to valid GGUF of format $f$, quality depends on the quantizer alone and is measured once, before coding.

\subsection{Entropy-aware quantization inside standard formats}\label{sec:quant}
\begin{figure}[t]
\centering
\includegraphics[width=\columnwidth]{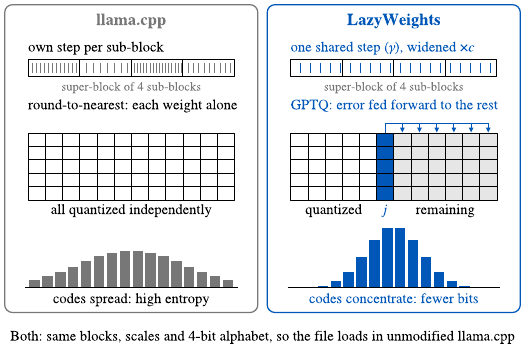}
\caption{Code selection (histograms illustrative). \sys{} keeps the format but shares one step per super-block ($\gamma$) and widens it ($c$), so the codes concentrate. GPTQ feeds each column's error to the remaining columns, so the coarser grid keeps its quality.}
\label{fig:ecsq}
\end{figure}
\sys{} keeps each format's blocks, scales and alphabet, so its files load in unmodified \lcpp{}, and changes only how the codes are chosen (Fig.~\ref{fig:ecsq}).

\textbf{Shared sub-block steps ($\gamma$).} \lcpp{} fits each sub-block of a 256-weight super-block to its own range, spreading its codes over the whole alphabet. This is right for fixed-width storage but wrong for a coder. We bound each sub-block's range from below by $\gamma$ times the largest range in its super-block. With $\gamma=0$ we recover \lcpp{}'s fit; with $\gamma=1$ the super-block shares one step, and its codes concentrate where its weights do.

\textbf{Step scale ($c$).} By~\eqref{eq:ent}, a wider step lowers the entropy. We therefore scale the fitted step by $c\in\{1,1.25,1.5\}$, chosen per operating point, and let GPTQ absorb the error.

\textbf{GPTQ error feedback.} The error feedback of GPTQ~\cite{gptq} makes the coarser grid affordable, whereas round-to-nearest (RTN) does not (\S\ref{sec:abl}). A weight $w$ in column $j$ takes the code
\begin{equation}\label{eq:round}
q^*=\arg\min_q\ h_{jj}^{-1}(w-\hat w_q)^2+\lambda\,\ell(q),
\end{equation}
where $h_{jj}^{-1}$ is the Hessian term, $\hat w_q$ the reconstruction and $\ell(q)$ the code length. All points use $\lambda=0$, since the ECSQ rate term~\cite{choi2017} does not pay off (\S\ref{sec:abl}). We calibrate on 128 sequences of 512 tokens from C4. The settings differ per point (e.g., $\gamma=1$, $c=1.5$ for Mistral-7B at $\le$3.5\%, but $\gamma=0$, $c=1.25$ for Phi-3 at $\le$1\%), which is why we solve~\eqref{eq:obj} by search rather than apply a fixed recipe.

\subsection{A lossless Huffman bundle built for parallel decoding}
\begin{figure*}[t]
\centering
\includegraphics[width=\textwidth]{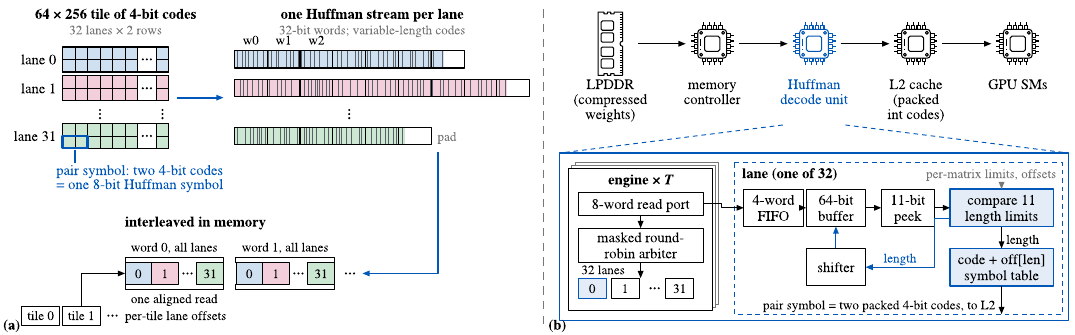}
\caption{(a) Bundle layout: a $64\times256$ tile of 4-bit codes is 32 lanes of two rows, each one Huffman stream of 32-bit words (code boundaries illustrative); interleaved words let one aligned read fetch the next word of every lane, and two 4-bit codes form one 8-bit pair symbol. (b) Decode unit between the memory controller and L2: $T$ engines of 32 lanes share an 8-word read port through a masked round-robin arbiter. A lane compares an 11-bit peek against per-length limits (no 2048-entry table), adds the length's offset to index the symbol table, and shifts its buffer by the length.}
\label{fig:arch}
\end{figure*}
Huffman decoding is sequential: a code's start is known only once the previous code is decoded. The bundle (Fig.~\ref{fig:arch}a) breaks this dependence. Each matrix has one canonical code, length-limited to 11 bits, so an 11-bit window always resolves a code. Each $64\times256$ tile is split into 32 lane streams of two rows, which decode in parallel without waiting for one another. The lanes' 32-bit words are interleaved, so the words needed next are adjacent, and a GPU warp or the unit's read port fetches them in wide, aligned reads; per-tile lane offsets let any tile start independently. In 4-bit formats, two codes form one 8-bit \emph{pair symbol}, so each decode step yields two weights.

\subsection{A Huffman decode unit between DRAM and L2}\label{sec:hw}
Decoding in the GPU kernels spends more time on bit manipulation than the saved bytes recover, and expanding weights back into DRAM saves storage, not bandwidth. \sys{} therefore decodes between the memory controller and L2 (Fig.~\ref{fig:arch}b): DRAM stores and moves only coded weights, L2 receives fixed-width codes, and the kernels are unchanged, so the first term of~\eqref{eq:time} follows the coded size.

\textbf{Lane and engine.} A lane feeds its stream through a 4-entry prefetch FIFO into a 64-bit bit buffer and resolves one code per cycle by comparing the next 11 bits against the 11 per-length limits at once (Fig.~\ref{fig:arch}b), emitting one pair symbol (two weights) per cycle; its engine broadcasts the per-matrix limit and offset tables. In an engine, 32 lanes share an 8-word read port through a masked round-robin arbiter and decode 31.9 symbols per cycle (0.2\% stall), i.e., 60--64 weights per cycle for pair-coded formats.

\textbf{Sizing and clock.}\label{sec:eng-clk} To keep pace with DRAM, the unit must decode
\begin{equation}\label{eq:theta}
\theta^*=\frac{8\beta}{b_c}
\end{equation}
weights per second, where $b_c$ is the coded bits per weight: 151--229\,G weights/s at 102\,GB/s. At 1\,GHz, close to the Orin's 1.02\,GHz GPU clock (MAXN\_SUPER), four engines (0.206\,mm$^2$) cover $\theta^*$ for five of the six points. SmolLM at $\le$1\%, whose 5- and 6-bit codes decode singly (32 weights per cycle per engine), needs five (0.258\,mm$^2$). Lower clocks need proportionally more engines, and the Nano needs 1--2. The lane's compare-and-shift recurrence closes 1\,GHz without the repair the arbiter needs, whereas a 2048-entry lookup table per lane is 9$\times$ larger and misses timing (Table~\ref{tab:synth}).

\textbf{Verification and prototype.} A C model reproduces every coded matrix of 17 bundles with zero mismatches; Verilator co-simulation of the SystemVerilog RTL is bit- and cycle-exact against it on 29/29 matrices in 5 configurations; and $N$ engines attached to a bandwidth-capped DRAM model with read latencies up to 1024 cycles decode all 6 points bit-exact at 95--96\% of $\theta^*$. We synthesize with Yosys/ABC, OpenROAD and OpenSTA on ASAP7 RVT TT (logic synthesis, pre-layout). For the FPGA prototype, four engines run on an AMD ZCU102 (Table~\ref{tab:platforms}), whose Arm cores load each bundle into block RAM in chunks and check every lane's output (CRC-32) and every chunk's cycle, stall and word counts against the C model.

\subsection{Runtime}\label{sec:runtime}
\begin{table}[t]
\caption{Runtime optimizations, before $\rightarrow$ after.}
\label{tab:runtime}
\centering\footnotesize
\setlength{\tabcolsep}{3pt}
\resizebox{\columnwidth}{!}{\begin{tabular}{@{}lll@{}}
\toprule
Change & Measured on & Effect \\
\midrule
Speculative attention, & SmolLM 8-token pass, depth 400 & 44.4 $\rightarrow$ 31.2\,ms \\
\quad GGUF-width twins & Phi-3 8-token pass, depth 900 & 77.5 $\rightarrow$ 52.5\,ms \\
Nano kernels & SmolLM $\le$1\% twin, greedy token & 474 $\rightarrow$ 152\,ms \\
 & SmolLM $\le$3.5\% & 282 $\rightarrow$ 134\,ms \\
 & Phi-3 & 585 $\rightarrow$ 261\,ms \\
Ordered reduction & Nano, 2-token coded pass & 418 $\rightarrow$ 301\,ms \\
CUDA graphs & Orin pass (485--645 kernels) & $-$1.9--3.5\,ms (7--12\%) \\
 & \quad per greedy token & $-$6--11\% \\
\bottomrule
\end{tabular}}
\end{table}
Our runtime executes both the coded bundle, decoded inside the GEMV kernels as the software reference, and its \emph{twin}: the same codes without entropy coding, as the GPU sees them behind the unit. A fixed-order reduction across tile columns, in the int8 and fp32 kernels, makes the two bit-identical, so we test losslessness end to end: logits are identical on both devices, and speculation (up to 8 tokens per pass) is token-identical to greedy decoding. Table~\ref{tab:runtime} lists the optimizations. For speculation, attention reads each key and value once for all draft tokens of a head, and the Q5\_K/Q6\_K twins keep the GGUF width (nibbles plus a high-bit plane) rather than bytes. On the instruction-bound Nano,\label{sec:eng-nano} a 32-bit fetch yields four (5/6-bit) or eight (4-bit) codes, byte permutes replace bit unpacking, integer-to-float conversion and table lookup without changing the multiply-add order, and the ordered reduction replaces Maxwell's slow global atomics. Each kernel launch idles the GPU for about 4\,$\mu$s, so we replay each pass shape (1--8 tokens) as a CUDA graph, with bit-identical logits on all 14 bundles; \lcpp{} already uses graphs on the Orin.

\textbf{Streaming on a 4\,GB device.}\label{sec:stream} The Nano has 3.6\,GB free at idle; CUDA takes about 200\,MB, and we keep 384\,MiB free because an out-of-memory kill takes down system services. Tegra device memory is pinned host memory and counts against this floor, so at most 2.56\,GB of weights stay resident, while the Mistral-7B bundles take 3.2--3.6\,GB. The rest is streamed in forward order every pass, read with O\_DIRECT by a loader thread into two pinned staging buffers that overlap I/O with compute. Streamed tokens are identical to resident ones.

\section{Experimental Setup}\label{sec:setup}
\textbf{Models and quality.} We evaluate Phi-3-mini-4k-instruct, Mistral-7B-Instruct-v0.2 and SmolLM-1.7B-Instruct on WikiText-2 perplexity (all tokens, 512-token chunks), zero-shot HellaSwag, ARC-easy and PIQA, and 8-shot chain-of-thought GSM8K with flexible extraction (the last number in the answer).

\textbf{Baselines.}\label{sec:eng-base} On the Orin, the budget baseline uses \lcpp{}'s CUDA kernels. The Nano is frozen at CUDA 10.2 and gcc 7.5 (its last JetPack), where current \lcpp{} no longer builds its CUDA backend and releases that do predate the I-quants (IQ4\_XS, IQ3\_M) several baselines use; its baseline is therefore current \lcpp{} cross-compiled for the 4-core CPU, the best \lcpp{} available. Our runtime builds for sm\_53 with fp32 activations. The \emph{fixed-width twin} isolates the unit's share with identical kernels; other runtimes (MLC-LLM, TensorRT-LLM) would alter only the kernel share.

\textbf{Latency and energy.} We generate 128 new tokens for each of 6 prompts (32 for streamed Mistral-7B on the Nano) and integrate energy from the module input rail, idle included, with Orin clocks pinned (MAXN\_SUPER). Every runtime runs greedy decoding (one weight pass per token) and \emph{prompt-lookup speculation}, which drafts up to 7 tokens from an $n$-gram table of the prompt and accepted text, with no drafter model, and verifies them in one pass. This yields 1.7--2.4 tokens per pass (0.41--0.58 passes per token). Each runtime uses its own drafter (\lcpp{}'s; ours, $n$=3); our runtime also ran a port of \lcpp{}'s to isolate the drafter's share (Fig.~\ref{fig:strip}a).

\textbf{Modeling the unit end to end.} Time with the unit is the twin's, measured in the same run and mode ($t_\text{hw}=t_\text{twin}$); this is conservative, since fewer DRAM bytes could also shorten the token, and the bandwidth-bound estimate scales $t_\text{twin}$ by the coded-to-fixed byte ratio. Energy is
\begin{equation}\label{eq:energy}
E_\text{hw}=E_\text{twin}-k\left(e_\text{DRAM}\,\Delta S-E_\text{dec,pass}\right),
\end{equation}
where $k$ is the number of weight passes per token ($1$ for greedy, measured under speculation) and $\Delta S$ the bytes saved per pass. $e_\text{DRAM}$, the input-power slope of a streaming DRAM read minus the same read from L2, is 90.5\,pJ/byte on the Orin and 34.9\,pJ/byte on the Nano. $E_\text{dec}$ uses synthesized engine power (\S\ref{sec:res-hw}), bounded by clock-gated and always-on engines with a 2$\times$ margin.

\section{Results}\label{sec:results}
\begin{table*}[t]
\caption{Operating points (the released GGUF files) and per-token cost on the Orin Nano Super. Upper row: the \texttt{llama.cpp} budget baseline (its format). b/w: all overheads; $\Delta$PPL vs fp16; GSM8K: flexible extraction; Saving $=1-b_\text{ours}/b_\text{llama.cpp}$. Orin: ms / mJ per token, each runtime with its own drafter; \sys{} with the unit at the measured twin time; $\Delta t$: bandwidth-bound / conservative; $\Delta E$: \eqref{eq:energy}, clock-gated decoder.}
\label{tab:headline}
\centering\footnotesize
\setlength{\tabcolsep}{3pt}
\resizebox{\textwidth}{!}{\begin{tabular}{@{}lll cc cccc c ccc ccc@{}}
\toprule
 & & & & & \multicolumn{4}{c}{Task accuracy} & & \multicolumn{3}{c}{Orin, greedy} & \multicolumn{3}{c}{Orin, speculation} \\
\cmidrule(lr){6-9}\cmidrule(lr){11-13}\cmidrule(l){14-16}
Model & Budget & Format & b/w & $\Delta$PPL & HellaSwag & ARC-e & PIQA & GSM8K & Saving & ms / mJ & $\Delta t$ & $\Delta E$ & ms / mJ & $\Delta t$ & $\Delta E$ \\
\midrule
Phi-3-mini & $\le$1\% & IQ4\_XS+im & 4.311 & +0.65\% & .783 & .807 & .798 & .785 &  & 35.0 / 691 &  &  & 27.4 / 495 &  &  \\
 &  & \sys{} & 4.046 & +0.11\% & .778 & .803 & .789 & .773 & 6.2\% & 31.4 / 627 & $-$15/$-$10\% & $-$9\% & 17.9 / 366 & $-$39/$-$35\% & $-$26\% \\
 & $\le$3.5\% & Q3\_K\_M+im & 4.108 & +2.91\% & .776 & .786 & .799 & .779 &  & 44.0 / 840 &  &  & 27.6 / 512 &  &  \\
 &  & \sys{} & 3.649 & +2.97\% & .774 & .811 & .796 & .770 & 11.2\% & 31.4 / 616 & $-$37/$-$29\% & $-$27\% & 18.3 / 370 & $-$42/$-$34\% & $-$28\% \\
\addlinespace[2pt]
Mistral-7B & $\le$1\% & Q5\_K\_M & 5.668 & +0.44\% & .842 & .716 & .768 & .425 &  & 83.3 / 1784 &  &  & 61.6 / 1276 &  &  \\
 &  & \sys{} & 3.886 & +0.94\% & .840 & .720 & .770 & .402 & 31.4\% & 53.0 / 1107 & $-$41/$-$36\% & $-$38\% & 41.5 / 871 & $-$37/$-$33\% & $-$32\% \\
 & $\le$3.5\% & IQ3\_M+im & 3.628 & +1.01\% & .838 & .722 & .779 & .400 &  & 77.7 / 1611 &  &  & 71.6 / 1397 &  &  \\
 &  & \sys{} & 3.422 & +1.46\% & .838 & .719 & .782 & .368 & 5.7\% & 52.9 / 1068 & $-$44/$-$32\% & $-$34\% & 38.5 / 786 & $-$56/$-$46\% & $-$44\% \\
\addlinespace[2pt]
SmolLM-1.7B & $\le$1\% & Q5\_K\_M & 5.721 & +0.65\% & .607 & .633 & .737 & .064 &  & 26.4 / 481 &  &  & 14.3 / 246 &  &  \\
 &  & \sys{} & 5.162 & +0.88\% & .607 & .630 & .739 & .067 & 9.8\% & 18.6 / 356 & $-$33/$-$29\% & $-$26\% & 10.1 / 194 & $-$33/$-$29\% & $-$21\% \\
 & $\le$3.5\% & IQ4\_XS+im & 4.388 & +3.09\% & .605 & .625 & .732 & .055 &  & 23.7 / 375 &  &  & 16.5 / 254 &  &  \\
 &  & \sys{} & 3.934 & +3.27\% & .601 & .610 & .736 & .055 & 10.3\% & 15.3 / 287 & $-$40/$-$36\% & $-$23\% & 7.6 / 143 & $-$57/$-$54\% & $-$44\% \\
\bottomrule
\end{tabular}}
\end{table*}

\subsection{Memory at equal quality}\label{sec:res-rate}
\begin{figure*}[t]
\centering
\includegraphics[width=\textwidth]{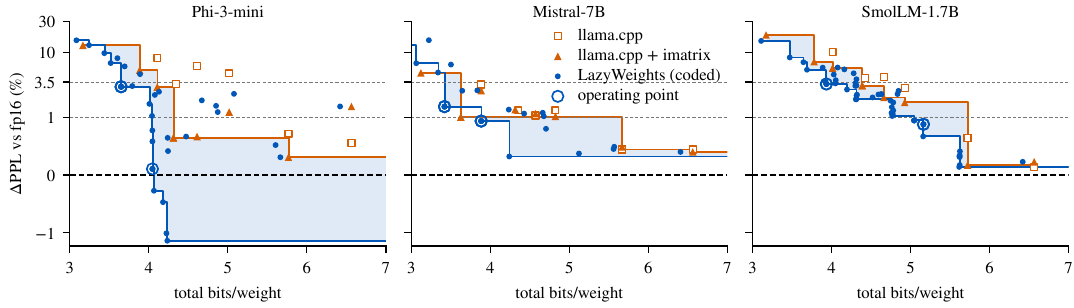}
\caption{Rate--quality frontiers. Dots: coded models, including \texttt{llama.cpp}'s imatrix codes behind our Huffman stage; lines: lowest $\Delta$PPL reachable at or below a rate; shading: where coding wins; rings: released operating points; dashed: the 1\% and 3.5\% budgets and fp16. Axis linear below 1\%.}
\label{fig:frontier}
\end{figure*}
Every coded model is 5.7--31.4\% smaller than the best \lcpp{} format within its budget, most of all Mistral-7B at $\le$1\% (Table~\ref{tab:headline}, Fig.~\ref{fig:frontier}). Huffman coding alone gives 6.2--20.7\% (coded vs.\ twin); the quantizer gives the rest, letting a coded file in a wider format undercut a narrower \lcpp{} format at equal quality. HellaSwag, ARC-easy and PIQA are at most 1.5 points below the baseline, and GSM8K at most 3.2 (Mistral at $\le$3.5\%).

\subsection{Latency and energy on the Orin Nano Super}\label{sec:res-e2e}
\begin{figure*}[t]
\centering
\includegraphics[width=\textwidth]{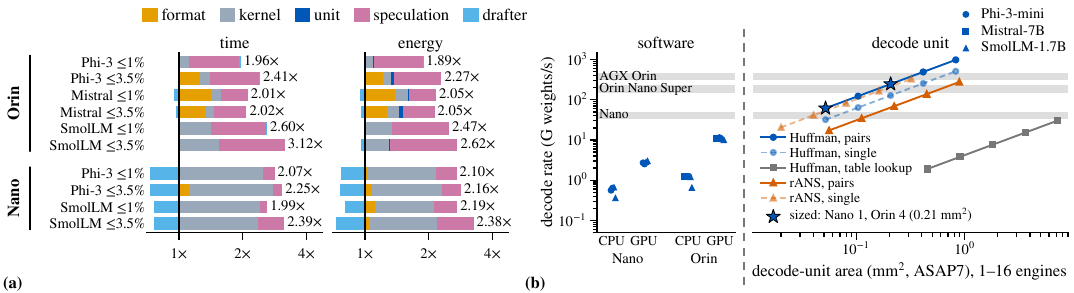}
\caption{(a) Factors from the \texttt{llama.cpp} baseline (greedy) to \sys{} + unit with our drafter, one change per step, on a log scale (steps below 1$\times$ extend left; unit time conservative; product at bar end). Top: Orin; bottom: Nano, resident points, where our drafter is a loss (0.73--0.76$\times$). (b) Decode rate. Left: software decoders on the coded operating points; Mistral-7B is streamed on the Nano. Right: area vs.\ rate for 1, 2, 4, 8 and 16 engines of 32 lanes (ASAP7 RVT TT, pre-layout) at the clock each closes; rANS engines share the Huffman arbiter; stars: the sized units. Bands: the rate DRAM delivers, $\theta^*=8\beta/b_c$ over the operating points; AGX Orin (not tested) at its 204.8\,GB/s specification.}
\label{fig:strip}
\end{figure*}

With the unit, every greedy point beats the budget baseline by 10--36\% in time (up to 44\% bandwidth-bound) and by 9--38\% in energy (Table~\ref{tab:headline}). The decoder costs 3--13\,mJ per token against 5--61\,mJ of DRAM energy saved; even with always-on engines at 2$\times$ power, energy still drops by 6--36\% greedy and 17--42\% with speculation. Without the unit, coding would slow the Orin down (\S\ref{sec:intro}, Fig.~\ref{fig:strip}b).

\textbf{Speculation composes with the unit.} Speculation cuts passes per token, and the unit cuts bytes per pass. Against \lcpp{} with speculation, every point is 29--54\% faster (up to 57\% bandwidth-bound) at 21--44\% less energy. Taking one change at a time (Fig.~\ref{fig:strip}a), speculation is the largest step on the Orin (1.3--2.0$\times$), the format gives 1.3--1.4$\times$ where it changes the file, and the drafter is within 5\%: ours accepts more tokens per pass (1.96 vs.\ 1.63 on Mistral), but larger passes cost more.

\subsection{The Jetson Nano}\label{sec:res-nano}
\begin{table}[t]
\caption{Jetson Nano, per token (ms / J; Mistral: s / J). CPU: \texttt{llama.cpp} with the budget baseline or our GGUF. GPU: our runtime, decoding in software or with the unit (twin time). Mistral streams from USB~3 at a 2560\,MB resident budget; its twin is no proxy for the unit (\S\ref{sec:disc}).}
\label{tab:nano}
\centering\footnotesize
\setlength{\tabcolsep}{2.5pt}
\resizebox{\columnwidth}{!}{\begin{tabular}{@{}ll cc cc@{}}
\toprule
 & & \multicolumn{2}{c}{CPU (\texttt{llama.cpp})} & \multicolumn{2}{c}{GPU (\sys{} runtime)} \\
\cmidrule(lr){3-4}\cmidrule(l){5-6}
Model & Budget & Baseline & \sys{} GGUF & Software & + unit \\
\midrule
\multicolumn{6}{@{}l}{\emph{Greedy}} \\
Phi-3 & $\le$1\% & 657 / 4.48 & 654 / 4.40 & 591 / 4.24 & 261 / 2.05 \\
 & $\le$3.5\% & 736 / 4.80 & 663 / 4.45 & 590 / 4.46 & 263 / 2.08 \\
SmolLM & $\le$1\% & 366 / 2.47 & 364 / 2.21 & 484 / 3.22 & 152 / 1.17 \\
 & $\le$3.5\% & 317 / 2.12 & 317 / 2.03 & 289 / 2.10 & 134 / 0.97 \\
Mistral & $\le$1\% & 22.6 / 71.2 & 18.0 / 58.3 & 3.40 / 15.8 & n/a \\
 & $\le$3.5\% & 1.93 / 12.7 & 15.1 / 49.0 & 2.08 / 12.0 & n/a \\
\midrule
\multicolumn{6}{@{}l}{\emph{Prompt-lookup speculation}} \\
Phi-3 & $\le$1\% & 840 / 5.60 & 865 / 5.74 & 423 / 2.94 & 244 / 1.71 \\
 & $\le$3.5\% & 956 / 6.23 & 864 / 5.74 & 415 / 3.05 & 239 / 1.69 \\
SmolLM & $\le$1\% & 475 / 3.10 & 437 / 2.66 & 286 / 1.84 & 141 / 0.91 \\
 & $\le$3.5\% & 419 / 2.73 & 469 / 2.94 & 170 / 1.17 & 101 / 0.65 \\
Mistral & $\le$1\% & 16.7 / 56.6 & 13.3 / 46.1 & 2.30 / 11.3 & n/a \\
 & $\le$3.5\% & 2.90 / 19.1 & 11.5 / 40.2 & 1.49 / 8.85 & n/a \\
\bottomrule
\end{tabular}}
\end{table}
For resident models (Table~\ref{tab:nano}), \sys{} with the unit takes 58--64\% less time and 53--57\% less energy per token than \lcpp{} on the CPU, greedy, and 70--76\% less time and 69--76\% less energy with speculation in both. Most of this comes from the GPU kernels (Fig.~\ref{fig:strip}a, bottom), as Table~\ref{tab:platforms} predicts, but the unit matters: software decoding is 2.2--3.2$\times$ slower than the twin and beats the CPU only on Phi-3 and SmolLM at $\le$3.5\% (by 9--20\%), not on SmolLM at $\le$1\%, whose 5/6-bit codes decode singly. With the Nano's low DRAM energy per byte, energy with the unit is within $\pm$0.03\,J of the measured twin energy. Speculation helps our GPU runtime (twin $-$25\% for SmolLM at $\le$3.5\%, but only $-$7 to $-$9\% for Phi-3, as an 8-token pass costs 3.6$\times$ a 1-token pass) and hurts \lcpp{} on the instruction-bound CPU by 28--32\%.

\textbf{Mistral-7B, streamed.} Here the coded size decides. At $\le$1\%, \sys{} is 6.6$\times$ faster than \lcpp{} at 4.5$\times$ less energy, even when decoding in software. At $\le$3.5\%, \lcpp{} narrowly wins greedy time at higher energy and \sys{} wins with speculation, because \lcpp{}'s memory-mapped IQ3\_M file (3.28\,GB) stays in the reclaimable page cache, which does not count against our floor. A slightly larger file gets almost no hits, as every token scans the weights in the same order: our IQ4\_XS GGUF (3.91\,GB) takes 15.1\,s. Our bundle (3.18\,GB) is smaller still, but the cache cannot hold its streamed matrices beside our pinned resident weights and the kernel's 464\,MB CMA reservation, hence O\_DIRECT. A pass costs about 1.0\,s of compute plus the streamed bytes, about the same for 1 or 8 tokens, so accepted tokens per pass set speculation's gain; a USB~3 stick (317\,MB/s) instead of the SD card (70\,MB/s) cuts our software-decoding token time 2.1--2.7$\times$.

\subsection{Decoder cost, and Huffman versus rANS}\label{sec:res-hw}
\begin{table}[t]
\caption{Decoder blocks, logic synthesis (pre-layout): ASAP7 RVT TT at 1\,GHz, Nangate45 at 500\,MHz.}
\label{tab:synth}
\centering\footnotesize
\setlength{\tabcolsep}{3pt}\setlength{\aboverulesep}{0pt}\setlength{\belowrulesep}{0pt}\renewcommand{\arraystretch}{1.1}
\begin{tabular*}{\columnwidth}{@{\extracolsep{\fill}}l rrr|rrr@{}}
\toprule
 & \multicolumn{3}{c|}{ASAP7} & \multicolumn{3}{c}{Nangate45} \\[-1pt]
Block & $\mu$m$^2$ & mW & slack & $\mu$m$^2$ & mW & slack \\
\midrule
Comparator lane & \hl{1,566} & \hl{3.59} & +117\,ps & 23,462 & 25.2 & +0.5\,ns \\
Table lane (2048 entries) & 14,076 & 94.4 & $-$31\,ns & -- & -- & -- \\
Arbiter, 32 lanes & \hl{1,443} & \hl{0.83} & met$^\dagger$ & 18,639 & 4.25 & +1.5\,ns \\
Arbiter, 64 lanes & 9,174 & -- & $-$0.29\,ns & -- & -- & -- \\
Engine & \hl{51,550} & \hl{115.7} & -- & -- & -- & -- \\
\bottomrule
\multicolumn{7}{@{}p{\columnwidth}@{}}{\raggedright$^\dagger$after OpenROAD repair. Power: co-simulation activity for the comparator lane, default otherwise. Shaded: Engine $=$ 32 comparator lanes $+$ arbiter.}
\end{tabular*}
\end{table}
A small unit clears the software decoding wall (Fig.~\ref{fig:strip}b; Table~\ref{tab:synth}).

\textbf{Huffman outperforms rANS on both axes.} In \emph{rate}, rANS is never smaller (0.0 to +1.0\% more bits): Huffman is already within 1.2--2.1\% of entropy, and rANS pays 32 bits of state plus padding per lane stream. In \emph{hardware}, at equal throughput (2\,G weights/s), rANS needs 4.0$\times$ the area with pair symbols (270\,MHz) or 1.15$\times$ with single codes (645\,MHz), because its one-cycle recurrence holds a symbol search, a table select and a 13$\times$20-bit multiply, against Huffman's compare and shift. Interleaving several rANS states per lane would relax its recurrence; we did not evaluate this.

\subsection{FPGA prototype}\label{sec:res-fpga}
On the ZCU102, four engines close timing at 112.5\,MHz (WNS +0.06\,ns; the decode-to-arbiter loop caps the design near 115\,MHz) and use 183.0k LUTs (67\%), 37.2k flip-flops, 464 BRAM36 (51\%, mostly payload copies) and no DSPs, i.e., 46k LUTs and 116 BRAM36 per engine. All 29 matrices (2,723 chunks, 484M weights) decode bit-exact, with cycle, stall and word counts equal to the C model on every chunk. Each engine sustains 31.8 weights per cycle, or 14.3\,G weights/s for four, and the fabric (VCCINT) draws 0.88\,W idle plus 3.7\,W at full rate, i.e., 0.26\,nJ per weight.

\section{Ablations}\label{sec:abl}
\begin{table}[t]
\caption{Ablations at the operating-point settings (b/w, $\Delta$PPL).}
\label{tab:abl}
\centering\footnotesize
\setlength{\tabcolsep}{2.5pt}
\resizebox{\columnwidth}{!}{\begin{tabular}{@{}llccc@{}}
\toprule
Model & Budget & Full & $\gamma$=0 & RTN \\
\midrule
Phi-3 & $\le$3.5\% & 3.649, +2.97 & 4.046, +0.11 & 3.646, +12.07 \\
Mistral & $\le$1\% & 3.886, +0.94 & 4.241, +0.32 & 3.886, $-$0.04 \\
 & $\le$3.5\% & 3.422, +1.46 & 3.837, +2.61 & 3.422, +2.78 \\
SmolLM & $\le$1\% & 5.162, +0.88 & 5.621, +0.31 & 5.162, +0.57 \\
 & $\le$3.5\% & 3.934, +3.27 & 4.321, +2.24 & 3.934, +6.14 \\
\bottomrule
\end{tabular}}
\end{table}
\textbf{Quantizer} (Table~\ref{tab:abl}). Shared sub-block steps account for most of the saving: with $\gamma=0$ (\lcpp{}'s fit), $\Delta$PPL is lower, but every point costs 9--12\% more bits. GPTQ feedback rescues the coarse points, which reach up to +12.1\% with RTN at the same bits, but not the fine ones, where RTN is better. Quality is non-monotone in the step scale (Mistral: +2.57\% at $\times$1.25, +1.46\% at $\times$1.5, +15.5\% at $\times$1.75), hence the per-model search. $\lambda>0$ saves at most 1.2\% of the bits and usually raises perplexity; int8 instead of fp32 activations moves perplexity by $-$0.07 to +0.18\%.

\textbf{Decoder.} Comparator vs.\ table lookup: Fig.~\ref{fig:strip}b. An 8-word read port with cross-stream prefetch reaches 31.9 symbols per cycle. Pair symbols change the size by $-$0.2 to +0.8\% but double an engine's rate. Reaching 99\% of the rate needs latency/4 FIFO words per lane (SRAM: +0.014\,mm$^2$ per engine). Wider engines only rescale the engine count: a 64-lane arbiter is 6.4$\times$ larger and misses 1\,GHz (Table~\ref{tab:synth}).

\textbf{Run-to-run variance.} Five more repeats of every measured Orin row of Table~\ref{tab:headline} vary by at most 0.6\% in time and energy per token (median 0.1\%), and every $\Delta t$ and $\Delta E$ stays within 1.7 points of the table.

\section{Limitations and Conclusion}\label{sec:disc}\label{sec:concl}
\textbf{Limitations.} The unit's latency and energy are projections from measured device numbers and pre-layout synthesis, not a fabricated SoC; the FPGA shows bit- and cycle-exact decoding, but at FPGA clocks rather than 1\,GHz. For streamed Mistral-7B, the twin is no proxy for the unit: it streams the fixed-width bytes (1.05\,GB per token at a 2560\,MB budget vs.\ 0.68/0.26\,GB coded at $\le$1\%/$\le$3.5\%), leaves less than our memory floor free, and at a 2048\,MB budget from the SD card was 1.2--1.8$\times$ slower than software decoding. We therefore do not report the unit's streaming time; reading its coded bytes alone takes 2.2/0.8\,s per token from USB. We did not evaluate Llama-3 or Qwen3, which quantize worse~\cite{llama3q}, or long contexts, where the uncoded fp16 KV cache dilutes the saving (Mistral-7B at 4K tokens: 0.54\,GB per pass vs.\ 3.5\,GB of weights).

\textbf{Conclusion.} The fixed width of edge LLM formats is a convention, not a necessity: the codes are far from uniform, the quantizer can make them less so at no cost in quality, and an entropy coder removes the difference if the decoder sits where bytes are counted. Shaped and Huffman-coded, models in \lcpp{}'s formats are 5.7--31.4\% smaller at equal perplexity, and a 0.2\,mm$^2$ unit between DRAM and L2 turns this into faster, cheaper tokens on a bandwidth-bound Orin Nano Super, with or without speculation, and lets a capacity-bound Nano run a 7B model that otherwise does not fit. Coding the KV cache, newer model families and a layout next to a real memory controller are next steps.

\section{Content Generated by AI}

In this work, we used generative AI tools (Claude, Anthropic) to implement methods. We have not used generative AI tools to generate synthetic data sets, to develop the theoretical models or conceptual frameworks underlying the method, to propose or refine hypotheses, to design or give feedback on research methodology or experiments, to support qualitative or thematic data analysis, to interpret results, to clean or reformat datasets, or for translation. The remaining tasks with required disclosure -- formulating mathematical claims, providing critical ingredients for proving mathematical claims, and assisting in the writing of proofs -- are not applicable to this work, which contains no formal proofs. Additionally, we used generative AI tools to create and edit software code, to create and modify scientific figures, to draft and edit parts of this paper, to identify relevant literature, to format references, and to search for information. We have reviewed all AI-assisted work.

Generative AI assistance was confined to implementation and exposition: the evaluation and ablation harness, the figure-generation scripts, and corrections to the quantization code. The authors validated all AI-assisted code and text against the experimental artifacts and take responsibility for the final content.

\section*{Acknowledgment}
We thank Radu Marculescu for lending us the AMD ZCU102 board, and Geffen Cooper for his help in setting it up.

\clearpage
\bibliographystyle{IEEEtran}
\bibliography{refs, swarm}

\end{document}